%% file: main.tex
\documentclass[conference]{IEEEtran}
\usepackage{times}

\usepackage[numbers]{natbib} 
\usepackage{multicol}
\usepackage{graphics,graphicx,caption,float,subcaption,booktabs,xcolor,multirow,array,color,ifthen,tabu,colortbl,dblfloatfix,url,xparse,mathtools,patchcmd,algorithmic,amssymb,xspace,nicefrac,microtype,amsmath,amsfonts,bm,ragged2e,tikz,stackengine,etoolbox,xpatch,enumerate,xstring,setspace,tabularx,makecell,changepage,cuted,titlesec,wrapfig,tcolorbox,footmisc, xspace}
\usepackage[pagebackref=true,breaklinks=true,colorlinks=true,bookmarks=false,citecolor=blue]{hyperref}
\usepackage[ruled,vlined]{algorithm2e}

\newcommand{\algo}{RMA}
\newcommand{\urllink}{https://youtu.be/p_pPrJXQk90}
\newcommand{\pagelink}{https://ashish-kmr.github.io/rma-legged-robots/}
\newcommand{\algofull}{Rapid Motor Adaptation}

\newcommand{\beginsupplement}{%
        \setcounter{table}{1}
        \renewcommand{\thetable}{S\arabic{table}}%
        \setcounter{figure}{1}
        \renewcommand{\thefigure}{S\arabic{figure}}%
        \setcounter{section}{1}
        \renewcommand{\thesection}{S\arabic{figure}}%
     }

\date{}
\begin{document}

\title{RMA: Rapid Motor Adaptation for Legged Robots}


\author{Ashish Kumar\\
UC Berkeley
\and
Zipeng Fu\\
Carnegie Mellon University
\and
Deepak Pathak\\
Carnegie Mellon University
\and
Jitendra Malik\\
UC Berkeley, Facebook
}



%

\makeatletter
\let\@oldmaketitle\@maketitle
\renewcommand{\@maketitle}{\@oldmaketitle
  \includegraphics[width=\linewidth,height=0.5\linewidth]{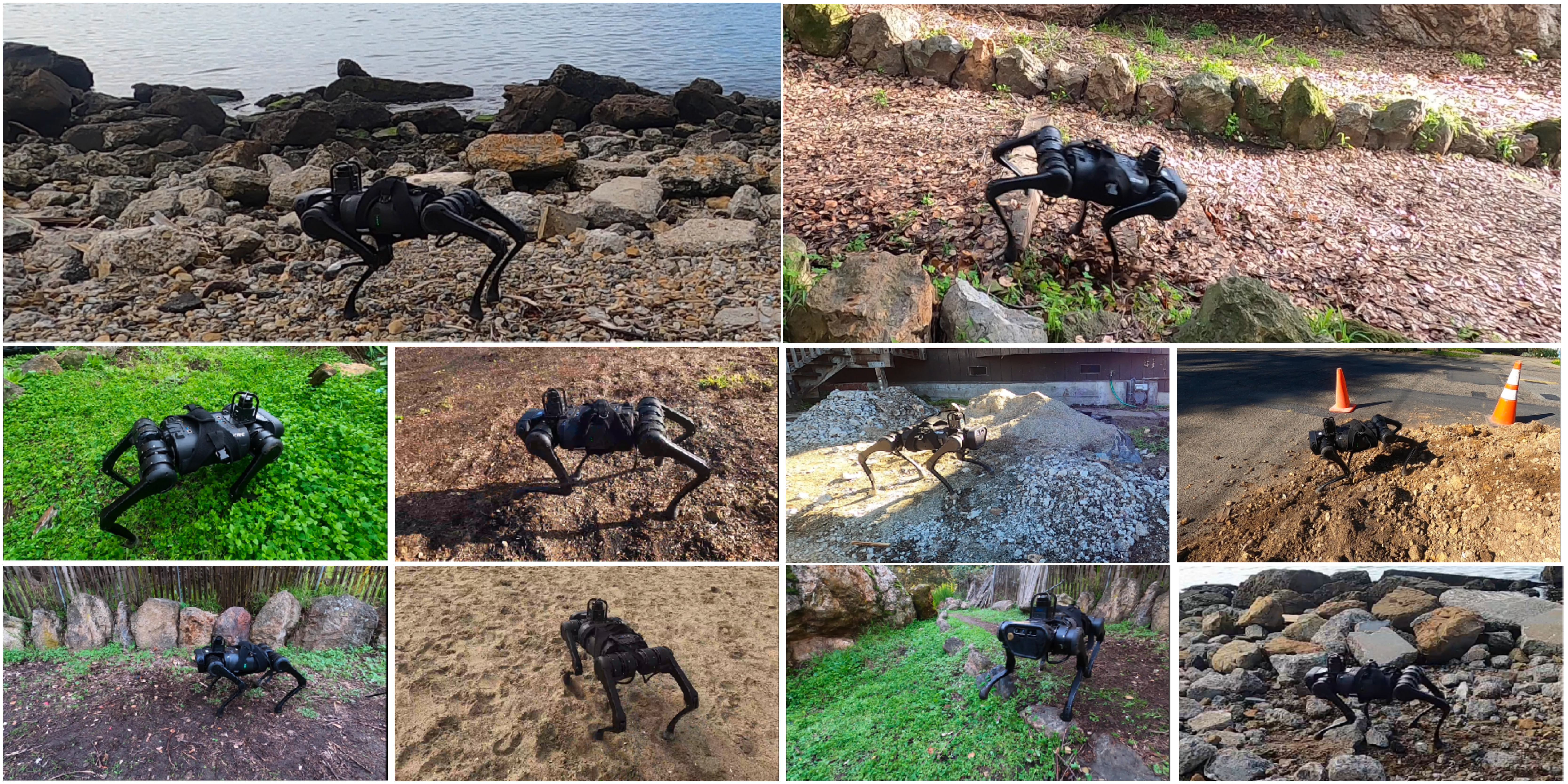}
  \captionof{figure}{We demonstrate the performance of \algo{} on several challenging environments. The robot is successfully able to walk on sand, mud, hiking trails, tall grass and dirt pile without a single failure in all our trials. The robot was successful in 70\% of the trials when walking down stairs along a hiking trail, and succeeded in 80\% of the trials when walking across a cement pile and a pile of pebbles. The robot achieves this high success rate despite never having seen unstable or sinking ground, obstructive vegetation or stairs during training. All deployment results are with the same policy without any simulation calibration, or real-world fine-tuning. Videos at~\url{\pagelink} 
  }
  \label{fig:outdoors}
  \bigskip}
\makeatother

\maketitle
\addtocounter{figure}{-1}


\begin{abstract}
Successful real-world deployment of legged robots would require them to \textit{adapt in real-time} to unseen scenarios like changing terrains, changing payloads, wear and tear. This paper presents Rapid Motor Adaptation (RMA) algorithm to solve this problem of real-time online adaptation in quadruped robots. \algo{} consists of two components: a base policy and an adaptation module. The combination of these components enables the robot to adapt to novel situations in fractions of a second. RMA is trained completely in simulation without using any domain knowledge like reference trajectories or predefined foot trajectory generators and is deployed on the A1 robot without any fine-tuning. We train RMA on a varied terrain generator using bioenergetics-inspired rewards and deploy it on a variety of difficult terrains including rocky, slippery, deformable surfaces in environments with grass, long vegetation, concrete, pebbles, stairs, sand, etc. RMA shows state-of-the-art performance across diverse real-world as well as simulation experiments. Video results at~\url{\pagelink}.
\end{abstract}

\IEEEpeerreviewmaketitle

\begin{figure*}
  \centering
  \includegraphics[width=0.9\linewidth]{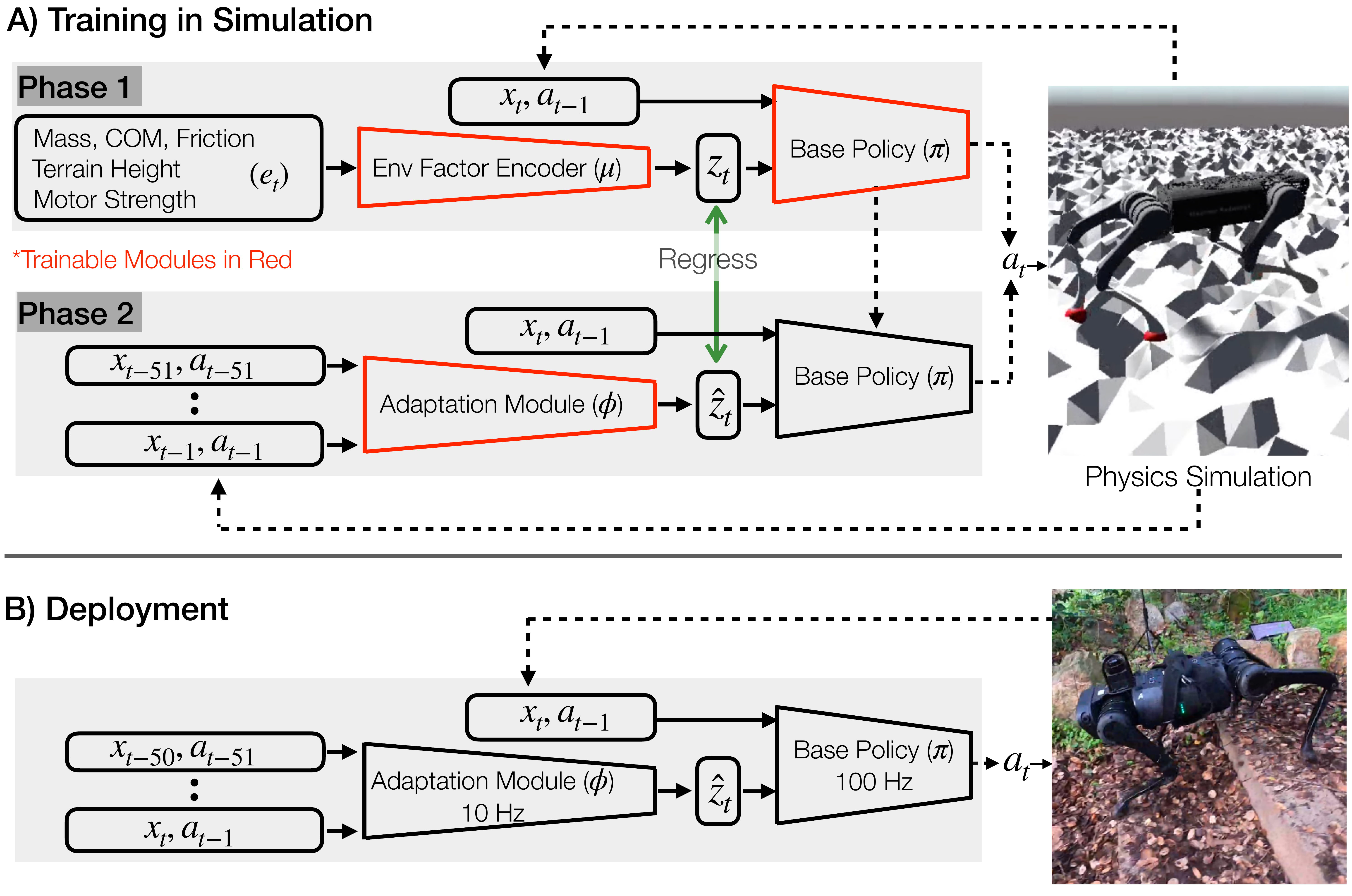}
\caption{
RMA consists of two subsystems - the base policy $\pi$ and the adaptation module $\phi$. \textbf{Top:} RMA is trained in two phases. In the first phase, the base policy $\pi$ takes as input the current state $x_t$, previous action $a_{t-1}$ and the privileged environmental factors $e_t$ which is encoded into the latent extrinsics vector $z_t$ using the environmental factor encoder $\mu$. The base policy is trained in simulation using model-free RL. In the second phase, the adaptation module $\phi$ is trained to predict the extrinsics $\hat{z_t}$ from the history of state and actions via supervised learning with on-policy data. \textbf{Bottom:} At deployment, the adaptation module $\phi$ generates the extrinsics $\hat{z}_t$ at 10Hz, and the base policy generates the desired joint positions at 100Hz which are converted to torques using A1's PD controller. Since the adaptation module runs at a lower frequency, the base policy consumes the most recent extrinsics vector $\hat{z}_t$ predicted by the adaptation module to predict $a_t$. This asynchronous design was critical for seamless deployment on low-cost robots like A1 with limited on-board compute. Videos at: \url{\pagelink}}
  \label{fig:method}
\end{figure*}

\section{Introduction}
\label{sec:introduction}
Great progress has been made in legged robotics over the last forty years through the modeling of physical dynamics and  the tools of control theory~\cite{miura1984dynamic,raibert1984hopping,saranli2001rhex,geyer2003positive,yin2007simbicon,zucker2010optimization,sreenath2011compliant,johnson2012tail,khoramshahi2013piecewise,ames2014rapidly,hyun2016implementation}. These methods require considerable expertise on the part of the human designer, and in recent years there has been much interest in replicating this success using reinforcement learning and imitation learning techniques~\cite{hwangbo2019learning,haarnoja2019learningtowalk,peng2020learning,yang2020data,lee2020learning} which could lower this burden, and perhaps also improve performance. The standard paradigm is to train an RL-based controller in a physics simulation environment and then transfer to the real world using various sim-to-real techniques~\cite{tobin2017domain,peng2018sim,hwangbo2019learning}.  This transfer has proven quite challenging, because the sim-to-real gap itself is the result of multiple factors: (a) the physical robot and its model in the simulator differ significantly; (b) real-world terrains vary considerably (Figure~\ref{fig:outdoors}) from our models of these in the simulator; (c) the physics simulator fails to accurately capture the physics of the real world – we are dealing here with contact forces, deformable surfaces and the like – a considerably harder problem than modeling rigid bodies moving in free space.

In this paper, we report on our progress on solving this challenge for quadruped locomotion, using as  an experimental platform the relatively cheap A1 robot from Unitree. Figure~\ref{fig:outdoors} shows some sample examples with in-action results in the video. Before outlining our approach (Figure~\ref{fig:method}), we begin by noting that human walking in the real world entails rapid adaptation as we move on different soils, uphill or downhill, carrying loads, with rested or tired muscles, and coping with sprained ankles and the like. Let us focus on this as a central problem for legged robots as well, and call it {\bf Rapid Motor Adaptation (RMA)}. We will posit that RMA has to occur online, at a time scale of fractions of a second, which implies that we have no time to carry out multiple experiments in the physical world, rolling out multiple trajectories and optimizing to estimate various system parameters.
It may be worse than that.  If we  introduce the quadruped onto a  rocky surface  with no prior experience, the robot policy would  fail often, causing serious damage to the robot. Collecting even 3-5 mins of walking data in order to adapt the walking policy may be practically infeasible. Our strategy therefore entails that not just the basic walking policy, but also RMA must be trained in simulation, and directly deployed in the real world. But, how?

Figure~\ref{fig:method} shows that \algo{} consists of two subsystems: the base policy $\pi$ and the adaptation module $\phi$, which work together to enable online real time adaptation on a very diverse set of environment configurations. 
The base policy is trained via reinforcement learning in simulation using privileged information about the environment configuration $e_t$ such as friction, payload, etc. Knowledge of the vector $e_t$ allows the base policy to appropriately adapt to the given environment. The environment configuration vector $e_t$ is first encoded into a latent feature space $z_t$ using an encoder network $\mu$. This latent vector $z_t$, which we call the \textit{extrinsics}, is then fed into the base policy along with the current state $x_t$ and the previous action $a_{t-1}$. The base policy then predicts the desired joint positions of the robot $a_t$. The policy $\pi$ and the environmental factor encoder $\mu$ are jointly trained via RL in simulation.

Unfortunately, this policy cannot be directly deployed because we don't have access to $e_t$ in the real world. What we need to do is to estimate the extrinsics at run time, which is the role of the adaptation module $\phi$. The key insight is that when we command a certain movement of the robot joints, the actual movement differs from that in a way that depends on the extrinsics. So instead of using privileged information, we might hope to use the recent history of the agent's state to estimate this extrinsics vector, analogously to the operation of a Kalman filter for state estimation from history of observables. Specifically, the goal of $\phi$  is to estimate the extrinsics vector $z_t$ from the robot's recent state and action history, without assuming any access to $e_t$. That is at runtime, but at training time, life is easier. {\bf Since both the state history and the extrinsics vector $\mathbf{z_t}$ can be computed in simulation, we can train this module via supervised learning}. At deployment, both these modules work together to perform robust and adaptive locomotion. In our experimental setup with its limited on-board computing, the base policy $\pi$ runs at 100 Hz, while the adaptation module $\phi$ is slower and runs at 10Hz. The two run asynchronously in parallel with no central clock  to align them. The base policy just uploads the most recent prediction of the  extrinsics vector $z_t$ from the adaptation module to predict action $a_t$.

Our approach is in contrast to previous learning-based work in locomotion that adapt learned policies via inferring the key parameters about the environment from a small dataset collected in every new situation to which the robot is introduced. These could either be physical parameters like friction, etc.~\cite{bongard2005nonlinear} or their latent encoding~\cite{peng2020learning}. Unfortunately, as mentioned earlier, collecting such a dataset, when the robot hasn't yet acquired a good policy for walking, could result in falls and damage to the robot. Our approach avoids this because RMA, through the rapid estimation of $z_t$ permits the walking policy to adapt quickly~\footnote{RMA takes less than $1$s, whereas~\citet{peng2020learning} need to collect $4-8$mins ($50$ episodes of $5-10$s) of data.} and avoid falls. 

Training of a base policy using RL with an extra argument for the environmental parameters has also been pursued in~\cite{yu2017preparing,peng2020learning}. Our novel aspects are the use of a varied terrain generator and “natural” reward functions motivated by bioenergetics which allows us to learn walking policies without using any reference demonstrations~\cite{peng2020learning}. But the truly novel contribution of this paper is the adaptation module, trained in simulation, which makes RMA possible. This, at deployment time, has the flavor of system identification, but it is an on-line version of system identification, based just on the single trajectory that the robot has seen in the past fraction of a second. 
One might reasonably ask why it should work at all, but we can offer a few speculations:
\begin{itemize}
\item System identification is traditionally thought of as an optimization problem. But in many settings researchers have found that given sample (input, output) pairs of optimization problems with their solutions, we could use a neural network to approximate the function mapping the problem to its solution~\cite{ahmed2009surrogate,guo2016convolutional}. Effectively that is what $\phi$ is learning to do. 
\item We don’t need perfect system identification for the approach to work.  The vector of extrinsics $z_t$ is a lower-dimensional nonlinear projection of the  environmental parameters. This takes care of some identifiability issues where some parameters could covary with identical effects on observables. Secondly, we don’t need this vector of extrinsics to be correct in some “ground truth” sense. What matters is that it leads to the “right” action, and the end-to-end training optimizes for that.
\item The range of situations seen in training should encompass
what the robot will encounter in the real world. We use a fractal terrain generator which accompanied by the randomization of parameters such as mass, friction etc. creates a wide variety of physical contexts in which the walking robot has to react. 
\end{itemize}

The most comparable work in terms of robust performance of RL policies for legged locomotion in the real-world is that of~\citet{lee2020learning} which, unlike our work, relies on hand-coded domain knowledge of predefined trajectory generator \cite{iscen2018policies} and motor models \cite{hwangbo2019learning}. We evaluated RMA across a wide variety of terrains in the real world (Figure~\ref{fig:outdoors}). The proposed adaptive controller is able to walk on slippery surfaces, uneven ground, deformable surfaces (such as foam, mattress, etc) and on rough terrain in natural environments such as grass, long vegetation, concrete, pebbles, rocky surfaces, sand, etc.  


\section{Related Work}
\label{sec:related}
Conventionally, legged locomotion has been approached by using control-based methods \cite{miura1984dynamic,raibert1984hopping,geyer2003positive,yin2007simbicon,sreenath2011compliant,johnson2012tail,khoramshahi2013piecewise,ames2014rapidly,hyun2016implementation,barragan2018minirhex}. MIT Cheetah 3 \cite{bledt2018cheetah} can achieve high speed and jump over obstacles by using regularized model predictive control (MPC) and simplified dynamics \cite{di2018dynamic}. The ANYmal robot \cite{hutter2016anymal} locomotes by optimizing a parameterized controller and planning based on an inverted pendulum model \cite{gehring2016practice}. However, these methods require accurate modeling of the real-world dynamics, in-depth prior knowledge of the robots, and manual tuning of gaits and behaviors. Optimizing controllers, combined with MPC, can mitigate some of the problems \cite{kober2013reinforcement,calandra2016bayesian,choromanski2018optimizing}, however they still require significant task-specific feature engineering \cite{de2010feature,gehring2016practice,apgar2018fast}. 

\vspace{0.6em}\noindent\textbf{Learning for Legged Locomotion}$\quad$ 
Some of the earliest attempts to incorporate learning into locomotion can be dated back to DARPA Learning Locomotion Program~\cite{zucker2010optimization,zucker2011optimization,ratliff2009chomp,zico2011stanford,kalakrishnan2010fast}. More recently, deep reinforcement learning (RL) offered an alternative to alleviate the reliance on human expertise and has shown good results in simulation~\cite{schulman2017proximal,lillicrap2016continuous,mnih2016asynchronous,fujimoto2018addressing}. However, such policies are difficult to transfer to the real world \cite{koos2010crossing,neunert2017off,boeing2012leveraging}. One approach is to directly train in the real world \cite{haarnoja2019learningtowalk,yang2020data}. However, such policies are limited to very simple setups, and scaling to complex setups requires unsafe exploration and a large number of samples. 

\vspace{0.6em}\noindent\textbf{Sim-to-Real Reinforcement Learning}$\quad$
To achieve complex walking behaviours in the real world using RL, several methods try to bridge the Sim-to-Real gap. Domain randomization is a class of methods in which the policy is trained with a wide range of environment parameters and sensor noises to learn behaviours which are robust in this range \cite{tan2018sim,tobin2017domain,peng2018sim,xie2020dynamics,nachum2020multi}. However, domain randomization trades optimality for robustness leading to an over conservative policy \cite{luo2017robust}.

Alternately, the Sim-to-Real gap can also be reduced by making the simulation more accurate \cite{hwangbo2019learning,tan2018sim,hanna2017grounded}. \citet{tan2018sim} improve the motor models by fitting a piece-wise linear function to data from the actual motors~\cite{tan2018sim}. \citet{hwangbo2019learning}, instead, use a neural network to parameterize the actuator model~\cite{hwangbo2019learning,lee2020learning}. However, these approaches require initial data collection from the robot to fit the motor model, and would require this to be done for every new setup. 

\vspace{0.6em}\noindent\textbf{System Identification and Adaptation}$\quad$
Instead of being agnostic to physics parameters, the policy can condition on these parameters via online system identification. During deployment in the real world, physics parameters can either be inferred through a module that is trained in simulation~\cite{yu2017preparing}, or be directly optimized for high returns by using evolutionary algorithms~\cite{yu2018policy}. Predicting the exact system parameters is often unnecessary and difficult, leading to poor performance in practice.
Instead, a low dimensional latent embedding can be used~\cite{peng2020learning,zhou2019environment}. At test time, this latent can be optimized using real-world rollouts by using policy gradient methods~\cite{peng2020learning}, Bayesian optimization~\cite{yu2019biped}, or random search~\cite{yu2020learning}. Another approach is to use meta learning to learn an initialization of policy network for fast online adaptation~\cite{finn2017model}. Although they have been demonstrated on real robots~\cite{song2020rapidly, clavera2018learning}, they still require multiple real-world rollouts to adapt. 


\section{Rapid Motor Adaptation}
\label{sec:method}
We now describe each component of the RMA algorithm introduced in the third paragraph of Section~\ref{sec:introduction} and summarized in Figure~\ref{fig:method}. Following sections discuss the base policy, the adaptation module and the deployment on the real-robot in order. We will use the same notation as introduced in Section~\ref{sec:introduction}.

\subsection{Base Policy}
We learn a base policy $\pi$ which takes as input the current state $x_t \in \mathbb{R}^{30}$, previous action $a_{t-1} \in \mathbb{R}^{12}$ and the extrinsics vector $z_t \in \mathbb{R}^{8}$ to predict the next action $a_t$. The predicted action $a_t$ is the desired joint position for the $12$ robot joints which is converted to torque using a PD controller. The extrinsics vector $z_t$ is a low dimensional encoding of the environment vector $e_t \in \mathbb{R}^{17}$ generated by $\mu$. 
\begin{align}
    z_t &= \mu(e_t)  \label{eq:mu}  \\
    a_t &= \pi(x_t, a_{t-1}, z_t) \label{eq:pi}
\end{align}

We implement $\mu$ and $\pi$ as MLPs (details in Section \ref{subsec:arch-details}). We jointly train the base policy $\pi$ and the environmental factor encoder $\mu$ end to end using model-free reinforcement learning. At time step $t$, $\pi$ takes the current state $x_t$, previous action $a_{t-1}$ and the extrinsics $z_t = \mu(e_t)$, to predict an action $a_t$. RL maximizes the following expected return of the policy $\pi$: 
\[
    J(\pi) = \mathbb{E}_{\tau \sim p(\tau|\pi)}\Bigg[\sum_{t=0}^{T-1}\gamma^t r_t\Bigg],
\]
where $\tau = \{(x_0, a_0, r_0), (x_1, a_1, r_1) . . .\}$ is the trajectory of the agent when executing policy $\pi$, and $p(\tau|\pi)$ represents the likelihood of the trajectory under $\pi$.

\vspace{0.6em}\noindent\textbf{Stable Gait through Natural Constraints:} Instead of adding artificial simulation noise, we train our agent under the following natural constraints. First, the reward function is motivated from bioenergetic constraints of minimizing work and ground impact~\cite{polet2019inelastic}. We found these reward functions to be critical for learning realistic gaits in simulation. Second, we train our policies on uneven terrain (Figure \ref{fig:method}) as a substitute for additional rewards used by~\cite{hwangbo2019learning} for foot clearance and robustness to external push. A walking policy trained under these natural constraints transfers to simple setups in the real world (like concrete or wooden floor) without any modifications. This is in contrast to other sim-to-real work which either calibrates the simulation with the real world~\cite{tan2018sim, hwangbo2019learning}, or fine-tunes the policy in the real world~\cite{peng2020learning}. The adaptation module then enables it to scale from simple setups to very challenging terrains as shown in Figure~\ref{fig:outdoors}.

\vspace{0.6em}\noindent\textbf{RL Rewards:} \label{subsubsec:rlrewards} The reward function encourages the agent to move forward with a maximum speed of 0.35 m/s, and penalizes it for jerky and inefficient motions. Let's denote the linear velocity as $\mathbf{v}$, the orientation as $\bm{\theta}$ and the angular velocity as $\bm{\omega}$, all in the robot's base frame. We additionally define the joint angles as $\mathbf{q}$, joint velocities as $\dot{\mathbf{q}}$, joint torques as $\bm{\tau}$, ground reaction forces at the feet as $\mathbf{f}$, velocity of the feet as $\mathbf{v_f}$ and the binary foot contact indicator vector as $\mathbf{g}$. The reward at time $t$ is defined as the sum of the following quantities:
\begin{enumerate}
    \item Forward: min$(v^t_x, 0.35)$ 
    \item Lateral Movement and Rotation: $- \|v^t_y\|^2 - \|\omega^t_{\texttt{yaw}}\|^2$
    \item Work: $-|\bm{\tau}^T \cdot (\mathbf{q}^t - \mathbf{q}^{t-1})|$
    \item Ground Impact: $-\|\mathbf{f}^t - \mathbf{f}^{t-1}\|^2$ 
    \item Smoothness: $-\| \bm{\tau}^{t} - \bm{\tau}^{t-1} \|^2$
    \item Action Magnitude: $- \|\mathbf{a}^t\|^2$
    \item Joint Speed: $- \|\dot{\mathbf{q}}^t\|^2$
    \item Orientation: $- \|\bm{\theta}^t_{\texttt{roll,\; pitch}}\|^2$
    \item Z Acceleration: $- \|v^t_z\|^2$
    \item Foot Slip:  $-\|\text{diag}(\mathbf{g}^t) \cdot \mathbf{v_f}^t\|^2$ 
\end{enumerate}
The scaling factor of each reward term is $20$, $21$, $0.002$, $0.02$, $0.001$, $0.07$, $0.002$, $1.5$, $2.0$, $0.8$ respectively. 

\vspace{0.6em}\noindent\textbf{Training Curriculum: } If we naively train our agent with the above reward function, it learns to stay in place because of the penalty terms on the movement of the joints. To prevent this collapse, we follow the strategy described in~\cite{hwangbo2019learning}. We start the training with very small penalty coefficients, and then gradually increase the strength of these coefficients using a fixed curriculum. We also linearly increase the difficulty of other perturbations such as mass, friction and motor strength as the training progresses. We don't have any curriculum on the terrains and start the training with randomly sampling the terrain profiles from the same fixed difficulty. 

\subsection{Adaptation Module}
The knowledge of privileged environment configuration $e_t$ and its encoded extrinsics vector $z_t$ are not accessible during deployment in the real-world. Hence, we propose to estimate the extrinsics online using the adaptation module $\phi$.
Instead of $e_t$, the adaptation module uses the recent history of robot's states $x_{t-k:t-1}$ and actions $a_{t-k:t-1}$ to generate $\hat{z_t}$ which is an estimate of the true extrinsics vector $z_t$. In our experiments, we use $k = 50$ which corresponds to $0.5$s. 
$$\hat{z_t} =  \phi\big(x_{t-k:t-1}, a_{t-k:t-1}\big)$$
Note that instead of predicting $e_t$, which is the case in typical system identification, we directly estimate the extrinsics $z_t$ that only encodes how the behavior should change to correct for the given environment vector $e_t$.

To train the adaptation module, we just need the state-action history and the target value of $z_t$ (given by the environmental factor encoder  $\mu$). Both of these are available in simulation, and hence, $\phi$ can be trained via supervised learning to minimize: $\text{MSE}(\hat{z_t}, z_t) = \| \hat{z_t} - z_t\|^2,$
where $z_t = \mu(e_t)$. We model $\phi$ as a $1$-D CNN to capture temporal correlations (Section \ref{subsec:arch-details}). 

One way to collect the state-action history is to unroll the trained base policy $\pi$ with the ground truth $z_t$. However, such a dataset will contain examples of only good trajectories where the robot walks seamlessly. Adaptation module $\phi$ trained on this data would not be robust to deviations from the expert trajectory, which will happen often during deployment. 

We resolve this problem by training $\phi$ with on-policy data (similar to~\citet{ross2011reduction}). We unroll the base policy $\pi$ with the $\hat{z_t}$ predicted by the randomly initialized policy $\phi$. We then use this state action history, paired with the \textit{ground truth} $z_t$ to train $\phi$. We iteratively repeat this until convergence. This training procedure ensures that \algo{} sees enough exploration trajectories during training due to \textit{a)} randomly initialized $\phi$, and \textit{b)} imperfect prediction of $\hat{z_t}$. This adds robustness to the performance of \algo{} during deployment. 

\subsection{Asynchronous Deployment}
We train \algo{} completely in simulation and then deploy it in the real world without any modification or fine-tuning. 
The two subsystems of \algo{} run asynchronously and at substantially different frequencies, and hence, can easily run using little on-board compute. The adaptation policy is slow because it operates on the state-action history of $50$ time steps, roughly updating the extrinsic vector $\hat{z_t}$ once every $0.1$s (10 Hz). The base policy runs at 100 Hz and uses the most recent $\hat{z_t}$ generated by the adaptation module, along with the current state and the previous action, to predict $a_t$. This asynchronous execution doesn't hurt performance in practice because $\hat{z_t}$ changes relatively infrequently in the real world.

Alternately, we could have trained a base policy which directly takes the state and action history as input without decoupling them into the two modules. We found that this (a) leads to unnatural gaits and poor performance in simulation, (b) can only run at 10Hz on the on-board compute, and (c) lacks the asynchronous design which is critical for a seamless deployment of \algo{} on the real robot without the need for any synchronization or calibration of the two subsystems. This asynchronous design is fundamentally enabled by the decoupling of the relatively infrequently changing extrinsics vector with the quickly changing robot state.

\begin{figure*}[t]
  \centering
  \includegraphics[width=0.95\linewidth]{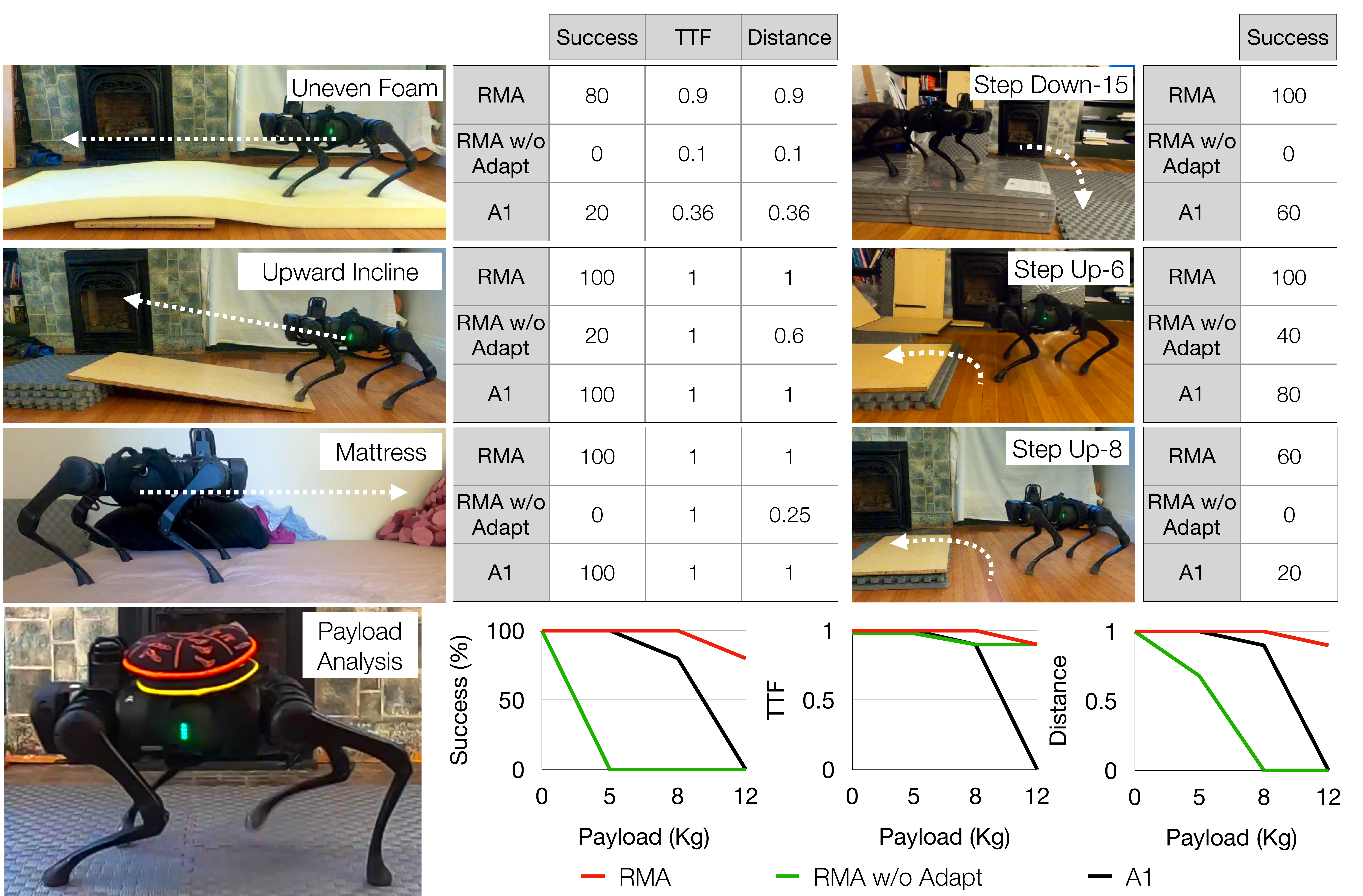}
  \caption{We evaluate \algo{} in several out-of-distribution setups in the real world. We compare \algo{} to A1's controller and \algo{} without the adaptation module. We find that \algo{} steps down a height of 15cm with 80\% success rate and walks over unseen deformable surfaces, such as a memory foam mattress and a slightly uneven foam with 100\% success rate. It is also able to successfully climb inclines and steps. A1's controller fails to walk over uneven foam. At the bottom, we also analyze the payload carrying limits of the three methods. We see that the A1 controller's performance starts degrading at 8Kg payload capacity. \algo{} w/o adaptation fails to move for payloads more than 8Kg, but rarely falls. For reference, A1 robot weights 12Kg. Overall, the proposed method consistently dominates the baseline methods. The numbers reported are averaged over 5 trials.}
  \label{fig:indoors}
  \vspace{-0.1in}
\end{figure*}

\begin{figure*}
  \centering
  \includegraphics[width=0.8\linewidth]{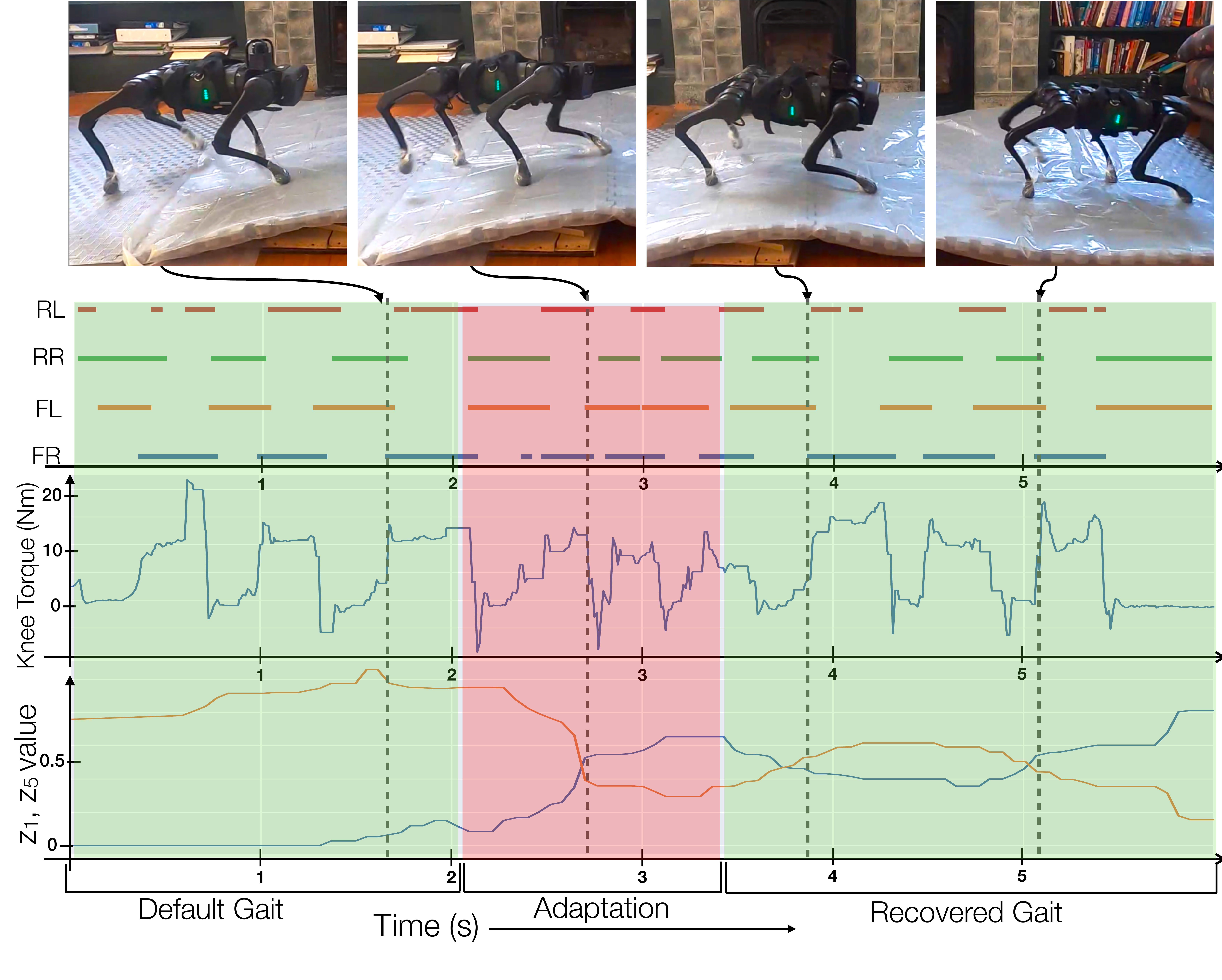}
  \vspace{-1em}
  \caption{We analyze \algo{} as the robot walks over an oily plastic sheet with additional plastic covering on its feet. We plot the torque of the knee and the gait pattern which indicates the contact of the four feet (F/R denotes Front/Rear and R/L denotes Right/Left). The bottom plot shows median filtered $1^{st}$ and $5^{th}$ components of the extrinsics vector $\hat{z}$ predicted by the adaptation module. When the robot enters the slippery patch we see a change in the two components of the extrinsics vector $\hat{z}$, indicating that the slip event has been detected by the adaptation module. Note that post adaptation, the recovered gait time period is similar to the original, the torque magnitudes have increased and $\hat{z}$ continues to capture the fact that the surface is still slippery. \algo{} was successful in 90\% of the runs over oily patch.}
  \label{fig:friction-analysis}
\end{figure*}


\section{Experimental Setup}
\label{sec:experiment}

\begin{table}[t]
\begin{center}
\begin{tabular}{lcc}
\toprule
 \textbf{Parameters} & \textbf{Training Range} & \textbf{Testing Range} \\
 \midrule
 Friction & [0.05, 4.5]  & [0.04, 6.0]  \\
 $K_p$ & [50, 60] & [45, 65]  \\
 $K_d$  &  [0.4, 0.8] & [0.3, 0.9] \\
 Payload (Kg)  & [0, 6]  & [0, 7]  \\
 Center of Mass (cm)  & [-0.15, 0.15]  & [-0.18, 0.18]  \\
 Motor Strength & [0.90, 1.10]  & [0.88, 1.22]  \\
 Re-sample Probability & 0.004 & 0.01 \\
\bottomrule
\end{tabular}
\vspace{-0.05in}
\caption{\label{tab:ood-range} Ranges of the environmental parameters.}
\end{center}
\vspace{-0.3in}
\end{table}

\subsection{Environment Details}
\label{subsec:env-details}
\noindent\textbf{Hardware Details:} We use A1 robot from Unitree for all our real-world experiments. A1 is a relatively low cost medium sized robotic quadruped dog. It has 18 degrees of freedom out of which 12 are actuated (3 motors on each leg) and weighs about 12 kg. To measure the current state of the robot, we use the joint position and velocity from the motor encoders, roll and pitch from the IMU sensor and the binarized foot contact indicators from the foot sensors. The deployed policy uses position control for the joints of the robots. The predicted desired joint positions are converted to torque using a PD controller with fixed gains ($K_p$ = $55$ and $K_d$ = $0.8$).

\vspace{0.6em}\noindent\textbf{Simulation Setup:} We use the RaiSim simulator \cite{raisim} for rigid-body and contact dynamics simulation. We import the A1 URDF file from Unitree \cite{unitree} and use the inbuilt fractal terrain generator to generate uneven terrain (fractal octaves = $2$, fractal lacunarity = $2.0$, fractal gain = $0.25$, z-scale = $0.27$). Each RL episode lasts for a maximum of $1000$ steps, with early termination if the height of the robots drops below $0.28$m, magnitude of the body roll exceeds $0.4$ radians or the pitch exceeds $0.2$ radians. The control frequency of the policy is $100$ Hz, and the simulation time step is $0.025$s.

\vspace{0.6em}\noindent\textbf{State-Action Space:} The state is $30$ dimensional containing the joint positions ($12$ values), joint velocities ($12$ values), roll and pitch of the torso and binary foot contact indicators ($4$ values). For actions,
we use position control for the $12$ robot joints. \algo{} predicts the desired joint angles $a = \hat{\mathbf{q}} \in \mathbb{R}^{12}$, which is converted to torques $\boldsymbol{\tau}$ using a PD controller: $\boldsymbol{\tau} = K_p \left(\hat{\mathbf{q}} - \mathbf{q}\right) + K_d \left(\hat{\dot{\mathbf{q}}} - \dot{\mathbf{q}}\right)$. $K_p$ and $K_d$ are manually-specified gains, and the target joint velocities $\hat{\dot{\mathbf{q}}}$ are set to 0. 

\vspace{0.6em}\noindent\textbf{Environmental Variations:} All environmental variations with their ranges are listed in Table \ref{tab:ood-range}. Of these, $e_t$ includes mass and its position on the robot (3 dims), motor strength (12 dims), friction (scalar) and local terrain height (scalar), making it a 17-dim vector. Note that although the difficulty of the terrain profile is fixed, the local terrain height changes as the agent moves. We discretize the terrain height under each foot to the first decimal place and then take the maximum among the four feet to get a scalar. This ensures that the controller does not critically depend on a fast and accurate sensing of the local terrain, and allows the base policy to use it asynchronously at a much lower update frequency during deployment.

\subsection{Training Details}
\label{subsec:arch-details}
\noindent\textbf{Base Policy and Environment Factor Encoder Architecture:} The base policy is a 3-layer multi-layer perceptron (MLP) which takes in the current state $x_t \in \mathbb{R}^{30}$, previous action $a_{t-1} \in \mathbb{R}^{12}$ and the extrinsics vector $z_t \in \mathbb{R}^{8}$, and outputs 12-dim target joint angles. The dimension of hidden layers is 128. The environment factor encoder is a 3-layer MLP (256, 128 hidden layer sizes) and encodes $e_t \in \mathbb{R}^{17}$ into $z_t \in \mathbb{R}^8$.

\vspace{0.6em}\noindent\textbf{Adaptation Module Architecture:} The adaptation module first embeds the recent states and actions into 32-dim representations using a 2-layer MLP. Then, a 3-layer $1$-D CNN convolves the representations across the time dimension to capture temporal correlations in the input. The input channel number, output channel number, kernel size, and stride of each layer are $[32, 32, 8, 4], [32, 32, 5, 1], [32, 32, 5, 1]$. The flattened CNN output is linearly projected to estimate $\hat{z_t}$. 

\vspace{0.6em}\noindent\textbf{Learning Base Policy and Environmental Factor Encoder Network:} We jointly train the base policy and the environment encoder network using PPO ~\cite{schulman2017proximal} for $15,000$ iterations each of which uses batch size of $80,000$ split into $4$ mini-batches. The learning rate is set to $5\mathrm{e}{-4}$. The coefficient of the reward terms are provided in Section \ref{sec:method}. Training takes roughly $24$ hours on an ordinary desktop machine, with 1 GPU for policy training. In this duration, it simulates $1.2$ billion steps. 

\vspace{0.6em}\noindent\textbf{Learning Adaptation Module:} We train the adaptation module using supervised learning with on-policy data. We use Adam optimizer ~\cite{kingma2014adam} to minimize MSE loss. We run the optimization process for $1000$ iterations with a learning rate of $5\mathrm{e}{-4}$ each of which uses a batch size of $80,000$ split up into $4$ mini-batches. It takes $3$ hours to train this on an ordinary desktop machine, with 1 GPU for training the policy. In this duration, it simulates $80$ million steps.

\section{Results and Analysis}
\label{sec:results}
We compare the performance of \algo{} with several baselines in simulation (Table \ref{tab:sim-metrics}). We additionally compare to the manufacturer's controller, which ships with A1, in the real world indoor setups (Figure \ref{fig:indoors}) and run \algo{} in the wild in a very diverse set of terrains (Figure \ref{fig:outdoors}). Videos at~\url{\pagelink}

\vspace{0.6em}\noindent\textbf{Baselines:} We compare to the following baselines:
\begin{enumerate}
    \item A1 Controller: The default robot manufacturer's controller which uses a force-based control scheme with MPC.
    \item Robustness through Domain Randomization (Robust): The base policy is trained without $z_t$ to be robust to the variations in the training range~\cite{tobin2017domain,peng2018sim}. 
    \item Expert Adaptation Policy (Expert): In simulation, we can use the true value of the extrinsics vector $z_t$. This is an upper bound to the performance of \algo{}. 
    \item \algo{} w/o Adaptation: We can also evaluate the performance of the base policy without the adaptation module to ablate the importance of the adaptation module.
    \item System Identification ~\cite{yu2017preparing}: Instead of predicting $\hat{z_t}$, we directly predict the system parameters $\hat{e_t}$. 
    \item Advantage Weighted Regression for Domain Adaptation (AWR) ~\cite{peng2020learning}: 
    Optimize $\hat{z_t}$ offline using AWR by using real-world rollouts of the policy in the testing environment.
\end{enumerate}
Learning baselines were trained with the same architecture, reward function and other hyper-parameters.  

\vspace{0.6em}\noindent\textbf{Metrics:}
We compare the performance of RMA against baselines using the following metrics:  (1) time-to-fall divided by maximum episode length to get a normalized value between $0-1$ (TTF); (2) average forward reward, (3) success rate, (4) distance covered, (5) exploration samples needed for adaptation, (6) torque applied, (7) smoothness which is derivative of torque and (7) ground impact (details in the supplementary).

\begin{table*}[t]
\begin{center}
\begin{tabular*}{0.95\textwidth}{@{}lcccccccc@{}}
\toprule
   & Success (\%) & TTF & Reward & Distance (m) & Samples & Torque & Smoothness & Ground Impact\\ 
\midrule
 Robust~\cite{tobin2017domain,peng2018sim} & 62.4 & 0.80 & 4.62 & 1.13 & 0 & 527.59 & 122.50 & 4.20\\
 SysID~\cite{yu2017preparing} & 56.5 & 0.74 & 4.82 & 1.17 & 0 & 565.85 & 149.75 & 4.03\\
 AWR~\cite{peng2020learning} & 41.7 & 0.65 & 4.17 & 0.95 & 40k & 599.71 & 162.60 & 4.02\\
 \algo{} w/o Adapt & 52.1 & 0.75 & 4.72 & 1.15 & 0 & 524.18 & 106.25 & 4.55\\
 \algo{} & 73.5 & 0.85 & 5.22 & 1.34 & 0 & 500.00 & 92.85 & 4.27\\
\midrule 
 Expert & 76.2 & 0.86 & 5.23 & 1.35 & 0 & 485.07 & 85.56 & 3.90\\
\bottomrule
\end{tabular*}
\caption{\label{tab:sim-metrics} \textbf{Simulation Testing Results:} We compare the performance of our method to baseline methods in simulation. Our train and test settings are listed in Table \ref{tab:ood-range}. We resample the environment parameters within an episode with a re-sampling probability of 0.01 per step during testing. Baselines and metrics are defined in Section \ref{sec:results}. The numbers reported are averaged over 3 randomly initialized policies and 1000 episodes per random initialization. \algo{} beats the performance of all the baselines, with only a slight degradation in performance compared to the Expert.
}
\end{center}
\vspace{-0.1in}
\end{table*}

\subsection{Indoor Experiments}
In the real world, we compare \algo{} with A1's controller and with \algo{} without the adaptation module (Figure \ref{fig:indoors}). We limit comparison to these two baselines to avoid damage to the robot hardware. We run 5 trials for each method and report the success rate, time to fall (TTF), and distance covered. Note that if a method drastically failed at a task, we only run two trials and then report a failure. This is done to minimize damage to the robot hardware. We have the following indoor setups:
\begin{itemize}
    \item \textbf{n-kg Payload}: Walk 300cm with n-kg payload on top.
    \item \textbf{StepUp-n}: Step up on an n-cm high step.
    \item \textbf{Uneven Foam}: Walk 180cm on a center elevated foam.
    \item \textbf{Mattress}: Walk 60cm on a memory foam mattress.
    \item \textbf{StepDown-n}: Step down an n-cm high step.
    \item \textbf{Incline}: Walk up on a 6-degrees incline.
    \item \textbf{Oily Surface}: Cross through an an oily patch.
\end{itemize}

Each trial of \textbf{StepUp-n} and \textbf{StepDown-n} is terminated after a success or a failure. Thus, we only report the success rate for these tasks because other metrics are meaningless.   

We observe that \algo{} achieves a high success rate in all these setups, beating the performance of A1's controller by a large margin in some cases. We find that turning off the adaptation module substantially degrades performance, implying that the adaptation module is critical to solve these tasks. A1's controller struggled with uneven foam and with a large step down and step up. The controller was destabilized by unstable footholds in most of its failures. In the payload analysis, the A1's controller was able to handle higher than the advertised payload (5Kg), but starts sagging, and eventually falls as the payload increases. In contrast, \algo{} maintains the height and is able to carry up to 12Kg (100\% of body weight) with a high success rate. \algo{} w/o adaptation mostly doesn't fall, but also doesn't move forward. We also evaluated \algo{} in a more challenging task of crossing an oily path with plastic wrapped feet. The robot successfully walks across the oily patch. Interestingly, \algo{} w/o adaptation was able to walk successfully on wooden floor without any fine-tuning or simulation calibration. This is in contrast to existing methods which calibrate the simulation ~\cite{tan2018sim, hwangbo2019learning} or fine-tune their policy at test time~\cite{peng2020learning} even for flat and static environments.

\subsection{Outdoor Experiments}
We demonstrate the performance of \algo{} on several challenging outdoor environments as shown in Figure \ref{fig:outdoors}. The robot is successfully able to walk on sand, mud and dirt without a single failure in all our trials. These terrains make locomotion difficult due to sinking and sticking feet, which requires the robot to change the footholds dynamically to ensure stability. \algo{} had a 100\% success rate for walking on tall vegetation or crossing a bush. Such terrains obstruct the feet of the robot, making it periodically unstable as it walks. To successfully walk in these setups, the robot has to stabilize against foot entanglements, and power through some of these obstructions aggressively. We also evaluate our robot on walking down some stairs found on a hiking trail. The robot was successful in 70\% of the trials, which is still remarkable given that the robot never sees a staircase during training. And lastly, we test the robot over construction debris, where it was successful 100\% of the times when walking downhill over a mud pile and 80\% of the times when walking across a cement pile and a pile of pebbles. The cement pile and pebbles were itself on a ground which was steeply sloping sideways, making it very challenging for the robot to go across the pile.

\subsection{Simulation Results}
We compare the performance of our method to baseline methods in simulation (Table~\ref{tab:sim-metrics}). We sample our training and testing parameters according to Table~\ref{tab:ood-range}, and resample them within an episode with a resampling probability of $0.004$ and $0.01$ per step respectively for training and testing. The numbers reported are averaged over 3 randomly initialized policies and 1000 episodes per random initialization. \algo{} performs the best with only a slight degradation compared to Expert's performance. The constantly changing environment leads to poor performance of AWR which is very slow to adapt. Since the Robust baseline is agnostic to extrinsics, it learns a very conservative policy which loses on performance. Note that the low performance of SysID implies that explicitly estimating $e_t$ is difficult and unnecessary to achieve superior performance. We also compare to \algo{} w/o adaptation, which shows a significant performance drop without the adaption module. 

\subsection{Adaptation Analysis}
We analyze the gait patterns, torque profiles and the estimated extrinsics vector $\hat{z_t}$ for adaptation over slippery surface (Figure \ref{fig:friction-analysis}). We pour oil on the plastic surface on the ground and additionally cover the feet of the robot in plastic. The robot then tries to cross the slippery patch and is able to successfully adapt to it. We found that \algo{} was successful in 90\% of the runs over oily patch. For one such trial, we plot the torque profile of the knee, the gait pattern, and median filtered $1^{st}$ and $5^{th}$ components of the extrinsics vector $\hat{z_t}$ in Figure \ref{fig:friction-analysis}. When the robot first starts slipping somewhere around $2$s, the slip disturbs the regular motion of the robot, after which it enters the adaptation phase. This is noticeable in the plotted components of the extrinsics vector which change in response to the slip. This detected slip enables the robot to recover and continue walking over the slippery patch. Note that although post adaptation, the torque stabilizes to a slightly higher magnitude and the gait time period is roughly recovered, the extrinsics vector does not recover and continues to capture the fact that the surface is slippery. See supplementary more such analysis. 

\section{Conclusion}
We presented the \algo{} algorithm for real-time adaptation of a legged robot walking in a variety of terrains. No demonstrations or predefined motion templates were needed. 
Despite only having access to proprioceptive data, the robot can also go downstairs and walk across rocks. However, a blind robot has limitations. Larger perturbations such as sudden falls while going downstairs, or due to multiple leg obstructions from rocks, sometimes lead to failures. To develop a truly reliable walking robot, we need to use not just proprioception but also  {\em exteroception} with an onboard vision sensor. The importance of vision in guiding long range, rapid locomotion has been well studied,  e.g. by ~\cite{matthis2018gaze}, and this is an important direction for future work.




\section*{Acknowledgments}
We would like to thank Jemin Hwangbo for helping with the simulation platform, and Koushil Sreenath and Stuart Anderson for helpful feedback during the course of this project. We would also like to thank Claire Tomlin, Shankar Sastry, Chris Atkeson, Aravind Sivakumar, Ilija Radosavovic and Russell Mendonca for their high quality feedback on the paper. This research was part of a BAIR-FAIR collaborative project, and recently also supported by the DARPA Machine Common Sense program.

\bibliographystyle{plainnat}
\bibliography{references}

\beginsupplement

\pagebreak
\twocolumn[{%
 \centering
 \LARGE Supplementary for \\
 RMA: Rapid Motor Adaptation for Legged Robots\\[1.5em]
}]
\input{supp-text}

\end{document}

%% file: supp-text.tex
\section{Metrics}
\label{sec:metrics}
We use several metrics (in SI units) to evaluate and compare the performance of RMA against baselines:
\begin{itemize}
    \item Success Rate: Average rate of successfully completing the task as defined in the next section. 
    \item Time to Fall (TTF): Measures the time before a fall. We divide it by the maximum duration of the episode and report a normalized value between $0$ and $1$.
    \item Reward: Average step forward reward plus lateral reward over multiple episodes as defined in Section III-A RL Rewards of the main paper. 
    \item Distance: Average distance covered in an episode. For real-world experiments, we report the normalized distance, where we normalize by the maximum distance which is specific to the task.
    \item Adaptation Samples: Number of control steps to explore in the testing environment needed for the motor policy to adapt.
    \item Torque: Squared L2 norm of torques at every joint $\| \bm{\tau}^{t} \|^2$.
    \item Jerk: Squared L2 norm of delta torques $\| \bm{\tau}^{t} - \bm{\tau}^{t-1} \|^2$.
    \item Ground Impact: Squared L2 norm of delta ground reaction forces at every foot $\|\mathbf{f}^t - \mathbf{f}^{t-1}\|^2$. 
\end{itemize}

\section{Additional Training and Deployment Details}
The training pipeline is shown in Algorithm \ref{algo:training}, and the deployment pipeline is shown in Algorithm \ref{algo:deployment}.

We use PPO \cite{schulman2017proximal} to train the base policy and the environmental factor encoder. We train for total $15,000$ iterations. During each iteration, we collect a batch of $80,000$ state-action transitions, which is evenly divided into $4$ mini-batches. Each mini-batch is fed into the base policy and the Environment Factor Encoder in sequence for 4 rounds to compute the loss and error back-propagation. The loss is the sum of surrogate policy loss and $0.5$ times the value loss. We clip the action log probability ratios between $0.8$ and $1.2$, and clip the target values to be within the $0.8 - 1.2$ times range of the corresponding old values. We exclude the entropy regularization of the base policy, but constrain the standard deviation of the parameterized Gaussian action space to be large than $0.2$ to ensure exploration. $\lambda$ and $\gamma$ in the generalization advantage estimation \cite{schulman2016gae} are set to $0.95$ and $0.998$ respectively. We use the Adam optimizer \cite{kingma2014adam}, where we set the learning rate to $5\mathrm{e}{-4}$, $\beta$ to $(0.9, 0.999)$, and $\epsilon$ to $1\mathrm{e}{-8}$. The reference implementation can be found in the RaisimGymTorch Library \cite{raisimgymtorch}. 

\begin{figure}[h]
  \centering
  \includegraphics[width=0.75\linewidth]{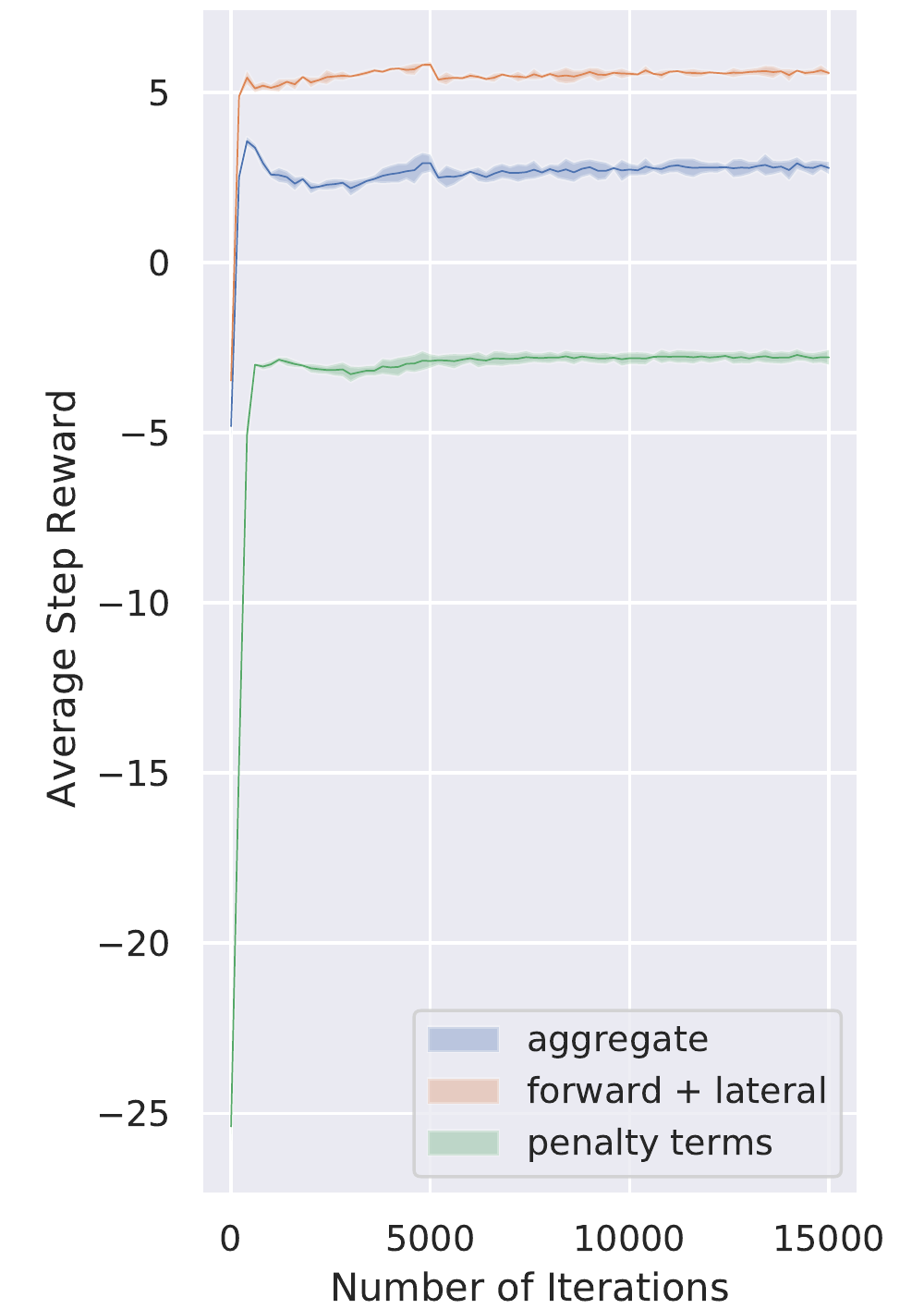}\\
  \caption{We plot the average step reward during the total $15,000$ training iterations. We show the converging trend of the reward aggregating all reward terms, forward + lateral reward, and sum of penalty terms. It also shows the necessity of applying a small multiplier to the penalty terms at the beginning of training; otherwise, the robot will only have negative experience initially and unable to learn to walk quickly.}
  \label{fig:training-reward}
\end{figure}

If we naively train our agent with the reward function aggregating all the terms, it learns to fall because of the penalty terms. To prevent this collapse, we follow the strategy described in~\cite{hwangbo2019learning}. In addition the scaling factors of all reward terms, we apply a small multiplier $k_t$ to the penalty terms $3-10$, as defined in Section III-A of the main paper. We start the training with a very small $k_0$ set to $0.03$, and then exponentially increase the these coefficients using a fixed curriculum: $k_{t+1} = k_{t}^{0.997}$, where $t$ is the iteration number. The learning process is shown in Figure \ref{fig:training-reward}. 

\begin{algorithm}[t]
\SetAlgoLined
 \textbf{\textit{Phase 1}} Randomly initialize the base policy $\pi$; Randomly initialize the environmental factor encoder $\mu$; Empty replay buffer $D_1$\;
 \For{$0 \leq \mathrm{itr} \leq N^{1}_{\mathrm{itr}}$}{
    \For{$0 \leq i \leq N_{\mathrm{env}}$}{
        $x_0, e_0 \gets$ envs[$i$].reset()\;
        \For{$0 \leq t \leq T$}{
            $z_t \gets \mu(e_t)$\;
            $a_t \gets \pi(x_t, a_{t-1}, z_t)$\;
            $x_{t+1}, e_{t+1}, r_t \gets $ envs[$i$].step($a_t$)\;
            Store $((x_t, e_t), a_t, r_t, (x_{t+1}, e_{t+1}))$ in $D_1$\;
        }
    }
    Update $\pi$ and $\mu$ using PPO \cite{schulman2017proximal}\;
    Empty $D_1$\;
 }
 \texttt{\\}
 \textbf{\textit{Phase 2}} Randomly initialize the adaptation module $\phi$ parameterized by $\theta_{\phi}$; Empty mini-batch $D_2$\;
 \For{$0 \leq \mathrm{itr} \leq N^{2}_{\mathrm{itr}}$}{
    \For{$0 \leq i \leq N_{\mathrm{env}}$}{
        $x_0, e_0 \gets$ envs[$i$].reset()\;
        \For{$0 \leq t \leq T$}{
            $\mathbf{\hat{z_t}} \gets \phi(x_{t-k:k}, a_{t-k-1:k-1})$\;
            $z_t \gets \mu(e_t)$\;
            $a_t \gets \pi(x_t, a_{t-1}, \mathbf{\hat{z_t}})$\;
            $x_{t+1}, e_{t+1}, \_ \gets $ envs[$i$].step($a_t$)\;
            Store $(\hat{z_t}, z_t)$ in $D_2$\;
        }
    }
    $\theta_{\phi} \gets \theta_{\phi} - \lambda_{\theta_{\phi}} \nabla_{\theta_{\phi}}\frac{1}{TN_{\mathrm{env}}}\sum{\| \hat{z_t} - z_t\|^2} $\;
    Empty $D_2$\;
 }

 \caption{\label{algo:training}\algofull Training}
\end{algorithm}

\begin{algorithm}[t]
\SetAlgoLined
\textbf{\textit{Process 1}} operating at 100 Hz\;
$t \gets 0$\;
\While{not fall}{
    $a_t \gets \pi(x_t, a_{t-1}, \hat{z}_{\mathrm{async}})$\;
    $x_{t+1} \gets $ env.step($a_t$)\;
    $t \gets t + 1$\;
}
\texttt{\\}
\textbf{\textit{Process 2}} operating at 10 Hz\;
\While{not fall}{
    $\hat{z}_{\mathrm{async}} \gets \phi(x_{t-k:k}, a_{t-k-1:k-1})$\;
}

 \caption{\label{algo:deployment}\algofull Deployment}
\end{algorithm}

\section{Additional Real-World Adaptation Anaylsis}
\begin{figure*}
  \centering
  \includegraphics[width=\linewidth]{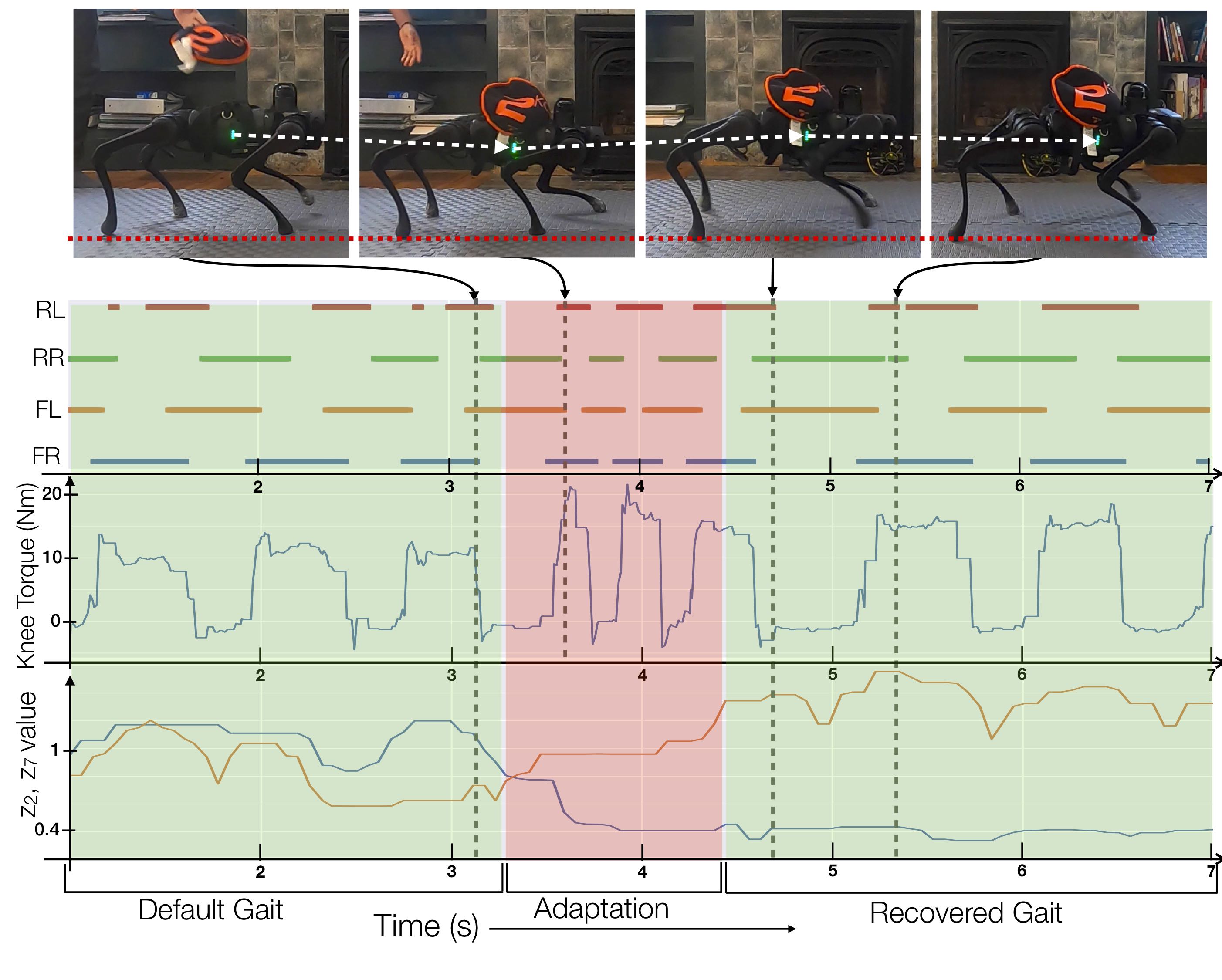}\\
  \caption{We analyze the change in behavior of \algo as we throw a payload of 5kg on the back of the robot. As a note, we have flipped the images so that that movement appears from left to right which is why the label on the sandbag appears to be 2Kg. We plot the torque profile of the knee and the gait pattern. The bottom plot shows median filtered $2^{nd}$ and $7^{th}$ components  of the extrinsics vector $\hat{z}$ predicted by the adaptation module. When the 5kg payload is thrown on the back of the robot, we see a dip in the center of mass of the robot, which the adaptation module subsequently recovers from. In the bottom plot, we see a jump in response in the plotted components of the estimated extrinsics vector, indicating that the additional payload has been detected by the adaptation module. Note that post adaptation, the recovered gait time period is roughly similar to the original, the torque magnitudes have increased and the extrinsics vector continues to capture the presence of the 5Kg payload on the back of the robot.}
  \label{fig:mass-analysis}
\end{figure*}

In addition to the oil-walking experiments in Figure 4 of the main paper, we also analyze the gait patterns and the torque profile for the mass adaptation case, shown in Figure \ref{fig:mass-analysis}. We throw a payload of 5kg on the back of the robot in the middle of a run and plot the torque profile of the knee, gait pattern, and the $2^{th}$ and $7^{th}$ components of the extrinsics vector $\hat{z_t}$ as shown in Figure \ref{fig:mass-analysis}. We observe that the additional payload disturbs the regular motion of the robot, after which it enters the adaptation phase and finally recovers from the disturbance. When the payload lands on the robot, it is noticeable that the plotted components of the extrinsics vector change in response to the slip. Post adaptation, we see that the torque stabilizes to a higher magnitude than before to account for the payload and the gait time period is roughly recovered.

\section{Additional Simulation Testings}
\begin{figure*}
  \centering
  \includegraphics[height=\linewidth]{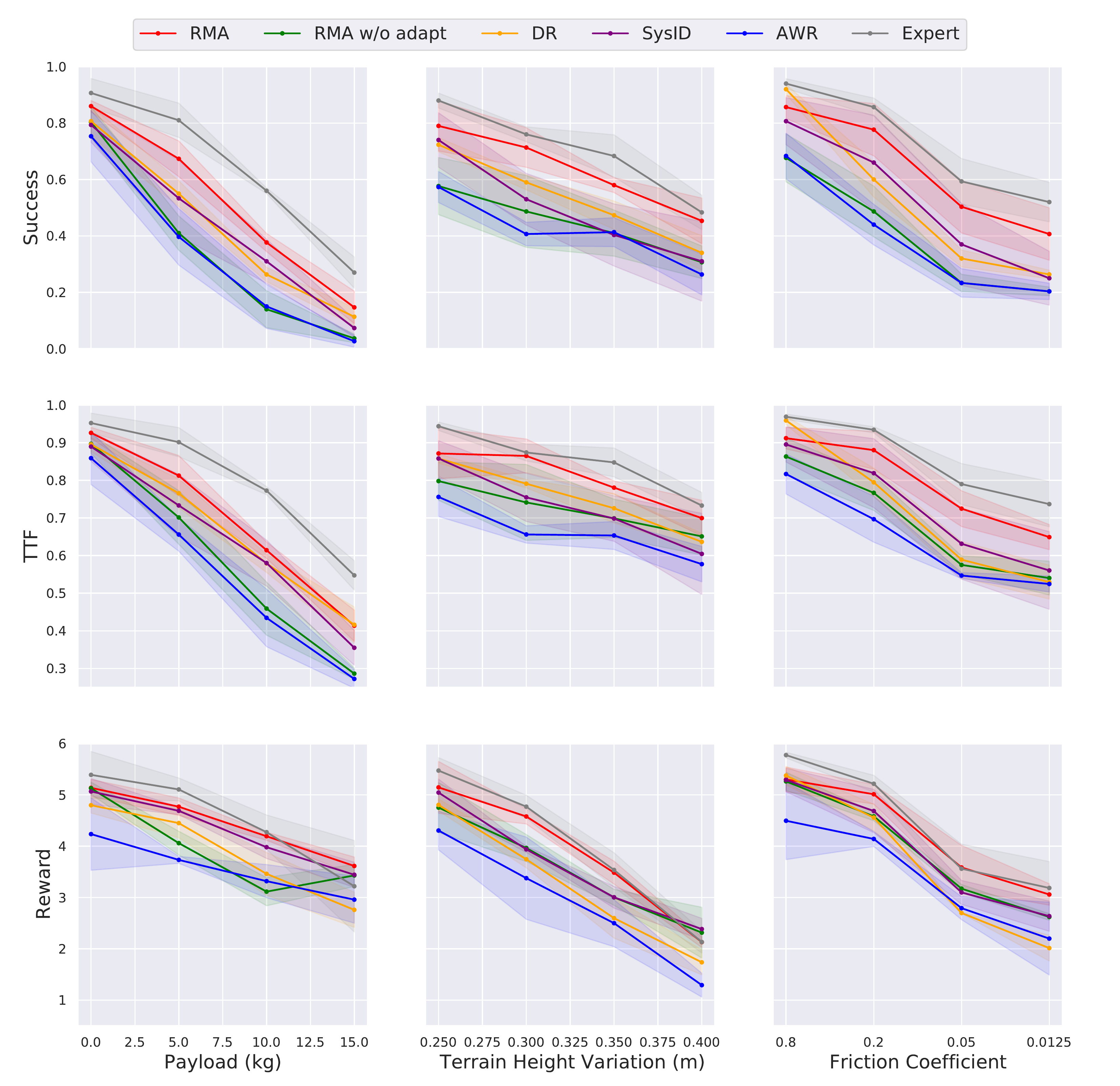}
  \caption{\textbf{Simulation Generalization Results:} We further compare the generalization performance of our method to baseline methods in simulation. We pick three physics parameters that may vary to a large degree in the real world: the payload on robot, the terrain height variation, and the friction coefficient between the robot feet and the terrain. We set other environment parameters according to the training range in TABLE II of the main paper. Baselines and metrics are defined in Section V of the main paper and Section \ref{sec:metrics}. We report the mean and standard deviation of the performance of 3 randomly initialized policies, which is characterized by the average of 100 testing trials in given settings. Despite no testing environment samples, \algo performs the best, the closest to Expert's performance. For reference, A1 robot without additional payloads weighs $12$ kg, and is $0.35$ m tall. The static friction coefficient between rubber and concrete is $1.00$. }
  \label{fig:generalization_test}
\end{figure*}

In Figure \ref{fig:generalization_test}, we further test \algo in extreme simulated environments and show its performance in three types of environment variations: the payloads added on the base of the A1 robot, the terrain elevation variation (z-scale used in the fractual terrain generator, details in Section IV Simulation Setup of the main paper), and the friction coefficient between the robot feet and the terrain. We show the superiority of \algo across all the cases in terms of Success Rate, TTF and Reward as defined in Section \ref{sec:metrics}.

